\documentclass[runningheads]{llncs}

\usepackage{eccv}

\usepackage{eccvabbrv}

\usepackage{graphicx}
\usepackage{booktabs}
\usepackage{amsmath}
\usepackage{amssymb}
\usepackage{multirow}
\usepackage{algorithm}
\usepackage{algorithmic}

\usepackage[accsupp]{axessibility}  

\usepackage[pagebackref,breaklinks,colorlinks,citecolor=eccvblue]{hyperref}

\usepackage{orcidlink}

\begin{document}

\title{Critic in the Loop: A Tri-System VLA Framework for Robust Long-Horizon Manipulation} 

\titlerunning{Critic in the Loop}

\author{Pengfei Yi \and Yingjie Ma \and Wenjiang Xu \and Yanan Hao \and Shuai Gan \and Wanting Li* \and Shanlin Zhong*
}

\authorrunning{Pengfei Yi et al.}

\institute{Institute of Automation, Chinese Academy of Sciences \and
the School of Artificial Intelligence, University of Chinese Academy of Sciences
\email{yipengfei2024@ia.ac.cn}\\
* co-corresponding authors
}
\maketitle

\begin{abstract}
  Balancing high-level semantic reasoning with low-level reactive control remains a core challenge in visual robotic manipulation. While Vision-Language Models (VLMs) excel at cognitive planning, their inference latency precludes real-time execution. Conversely, fast Vision-Language-Action (VLA) models often lack the semantic depth required for complex, long-horizon tasks. To bridge this gap, we introduce Critic in the Loop, an adaptive hierarchical framework driven by dynamic VLM-Expert scheduling. At its core is a bionic Tri-System architecture comprising a VLM brain for global reasoning, a VLA cerebellum for reactive execution, and a lightweight visual Critic. By continuously monitoring the workspace, the Critic dynamically routes control authority. It sustains rapid closed-loop execution via the VLA for routine subtasks, and adaptively triggers the VLM for replanning upon detecting execution anomalies such as task stagnation or failures. Furthermore, our architecture seamlessly integrates human-inspired rules to intuitively break infinite retry loops. This visually-grounded scheduling minimizes expensive VLM queries, while substantially enhancing system robustness and autonomy in out-of-distribution (OOD) scenarios. Comprehensive experiments on challenging, long-horizon manipulation benchmarks reveal that our approach achieves state-of-the-art performance.
\end{abstract}

\section{Introduction}
\label{sec:intro}

\begin{figure}
\begin{center}
\includegraphics[width=\textwidth]{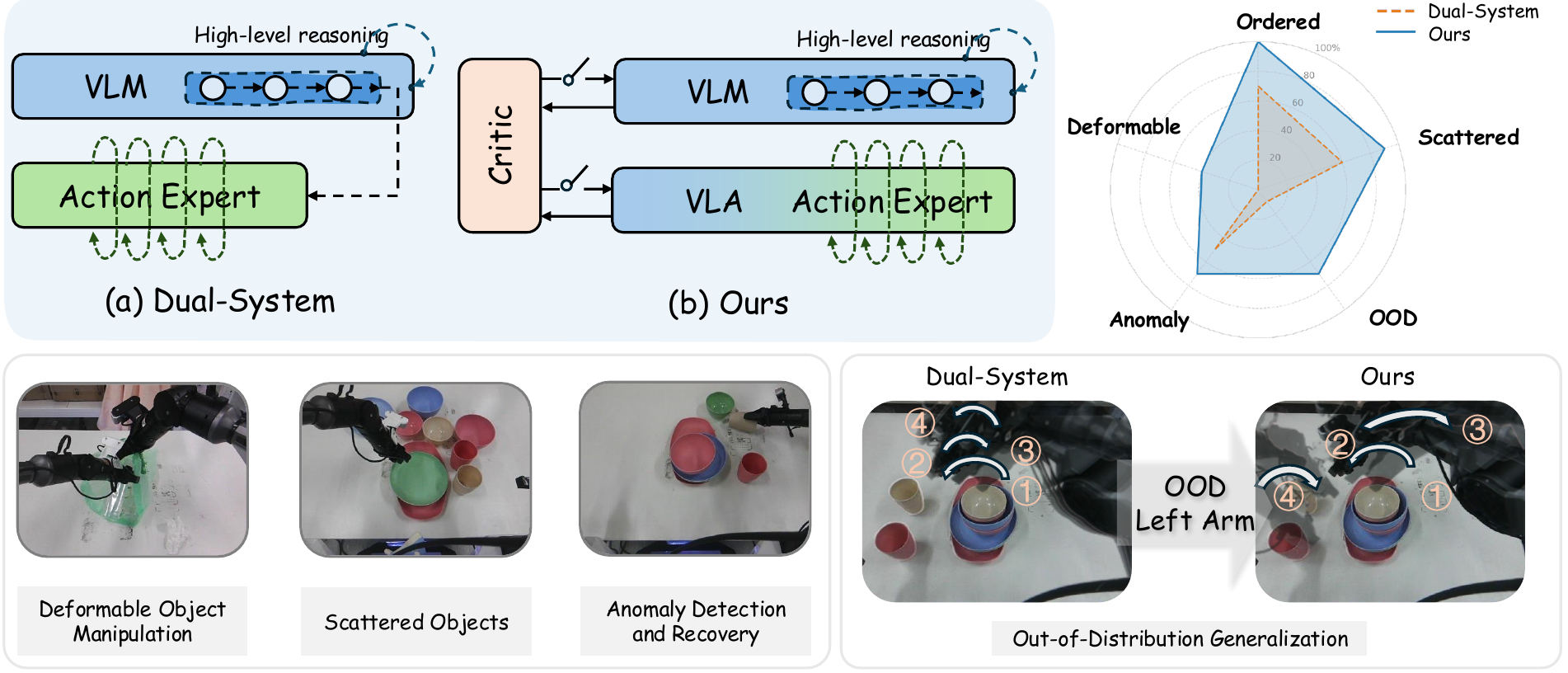}
\end{center}
   \caption{Overview. (a) Previous static dual-system pipeline. (b) Ours dynamically routes between a high-level VLM and VLA via an independent Critic. The right radar chart highlights our superior success rates over the baseline across diverse scenes. The bottom panels showcase real-world capabilities, notably demonstrating out-of-distribution (OOD) generalization where our system successfully picks and places a cup using an OOD left arm, despite lacking left-arm training data for this task. }
\label{fig:intro}
\end{figure}

A distinctive hallmark of human intelligence is the seamless integration of high-level planning with fine-grained physical execution\cite{varela1991embodied,varela2017embodied}. When cleaning a cluttered room, for example, we reason about a sequence of sub-goals (unfold the trash bag, open it, place items inside) while simultaneously converting those goals into precise motor actions. These processes are tightly coupled: reasoning steers action, and immediate physical feedback reshapes subsequent reasoning. Our goal is to endow robots with a similarly flexible, synergistic interplay between deliberation and control in real-world manipulation.

Many Vision-Language-Action (VLA) approaches\cite{pi0.5,han2024dual,shi2025hirobot,chen2025fast,cui2025openhelix,li2025hamster,lin2025onetwovla,bjorck2025gr00t} adopt Kahneman’s dual-system metaphor\cite{kahneman2011thinking}: a slow System Two (e.g., internet-pretrained VLMs\cite{beyer2024paligemma}) proposes sub-goals, while a fast System One (e.g., VLA policies\cite{kim2024openvla}) executes low-level commands. This design, however, has persistent bottlenecks. The switching policy is often rigid (fixed-rate or heuristic), wasting compute during smooth execution and reacting sluggishly to disturbances\cite{cui2025openhelix}.
Second, handling rare but critical failures typically requires costly, task-specific data collection, which limits scalability to long-horizon tasks\cite{li2025hamster,lin2025onetwovla}.

We argue that physical adaptability requires an architecture that explicitly knows \textit{when} to think, preserving mutual awareness between planning and execution. We therefore introduce the Tri-System VLA, an asynchronous architecture that decouples cognitive reasoning from continuous control via event-driven scheduling. Our framework introduces a third pillar: System Three (The Critic). Unlike traditional binary failure classifiers, our Critic provides continuous progress tracking and discrete anomaly detection. This allows the system to remain in a high-frequency ``Acting Mode'' (System One) for reactive control, awakening the ``Brain'' (System Two) only upon subtask completion, physical failure, or detected stagnation. Furthermore, this architecture inherently facilitates the integration of human-inspired rules. Specifically, when stagnation is detected, the system triggers a heuristic-guided state reset. By leveraging the Critic's insights to break infinite retry loops, this approach significantly enhances the system's ability to handle out-of-distribution (OOD) scenarios without requiring exhaustive emergency-scenario data.

To alleviate the reliance on expensive human-annotated subtask data, we develop an automated subtask annotation pipeline that segments demonstrations and extracts semantic labels. This pipeline enables robust learning from datasets without manual effort.

Our experiments validate the Tri-System VLA and demonstrate that it effectively addresses key bottlenecks in embodied intelligence. Our main contributions are threefold:
\begin{itemize}
    \item \textbf{Adaptive Cognitive Switching:} We introduce a critic-guided asynchronous scheduling mechanism that dynamically invokes high-level reasoning, drastically improving computational efficiency and physical responsiveness.
    \item \textbf{Proactive Anomaly Detection and Recovery:} We seamlessly integrate state-recovery mechanisms driven by a combination of human-inspired rules and data-backed strategies. This comprehensive detection intuitively breaks infinite retry loops, substantially enhancing system robustness and autonomy in out-of-distribution scenarios.
    \item \textbf{Scalable Subtask Annotation Pipeline:} We develop an automated subtask extraction tool that eliminates the manual data bottleneck, enabling robust long-horizon training from diverse datasets.
\end{itemize}

\section{Related Work}
\label{sec:related_work}

\subsection{Hierarchical Vision-Language-Action Models} 

Recent research has pivoted towards hierarchical and bi-level Vision-Language-Action (VLA) architectures to handle long-horizon reasoning and complex open-world manipulation. To bridge the gap between abstract semantics and continuous control, $\pi_{0.5}$ \cite{pi0.5} introduces a hierarchical structure employing semantic subtask prediction to guide low-level flow-matching experts, whereas HAMSTER \cite{li2025hamster} employs 2D spatial paths as intermediate guidance for 3D-aware policies, and VAMOS \cite{castro2025vamos} decouples semantic planning from embodiment grounding. Frameworks like OneTwoVLA \cite{lin2025onetwovla} and Hi Robot \cite{shi2025hirobot} utilize dual-system and hierarchical approaches that decouple high-level explicit reasoning from low-level action execution, enabling adaptive mode-switching and dynamic instruction following. However, while decoupling reasoning and control is advantageous, relying on fixed-frequency switching or inherent sequential orders between hierarchical systems often induces execution rigidity, severely limiting task performance.
\subsection{Robot Task Value Estimation} 

Recent research in embodied AI predominantly restricts Value Estimation Models to serving merely as underlying reward functions for reinforcement learning. For instance, within this paradigm, Robo-Dopamine\cite{tan2025robo} and RoboReward\cite{lee2026roboreward} construct a process evaluation mechanism to measure the fine-grained value contribution of each intermediate step in an execution trajectory to the overall task; GR-RL\cite{li2025gr} further utilizes generative models to directly perform structured value reasoning and evaluation regarding the current completion status of complex tasks. Building upon this, RISE\cite{yang2026rise} performs value estimation within generative world models, projecting the future evolution of current states to measure their value to the overall task, thereby bypassing current physical observation limits. In terms of physical deployments, $\pi^*_{0.6}$ \cite{intelligence2025pi} directly applies value estimation to heterogeneous online data streams, steering the progression of complex tasks through continuous value judgments of execution states. However, confined to this evaluative role, existing methods still struggle to react promptly to sudden online perturbations and lack a systematic framework to holistically evaluate overall task completion.

\subsection{Automated Subtask Annotation} 

With the rapid advancement of large Vision-Language Models (VLMs), recent research has increasingly focused on the automatic annotation of key states in Vision-Language-Action (VLA) data to facilitate model training. For instance, FoundationMotion \cite{gan2025foundationmotion} couples object tracking with Large Language Models (LLMs) to synthesize structured motion trajectories and enable spatial reasoning. In the domain of visually-guided robotic manipulation, video2tasks \cite{video2tasks_2025} automates task boundary detection, thereby isolating the structural milestones necessary for training reward models such as Robo-Dopamine \cite{tan2025robo}. Furthermore, Logic-in-Frames \cite{guo2025logic} advances this direction by formulating the milestone selection process as an iterative semantic-logical search. Similarly, K-frames \cite{yao2025kframes} addresses this extraction process through a scene-driven reinforcement learning objective. Crucially, however, while these methods can successfully abstract key states from dense video streams, their fundamental lack of explicit physical space constraints makes them highly susceptible to severe hallucinations in real-world physical deployments.

\begin{figure}[t]
\begin{center}
\includegraphics[width=\textwidth]{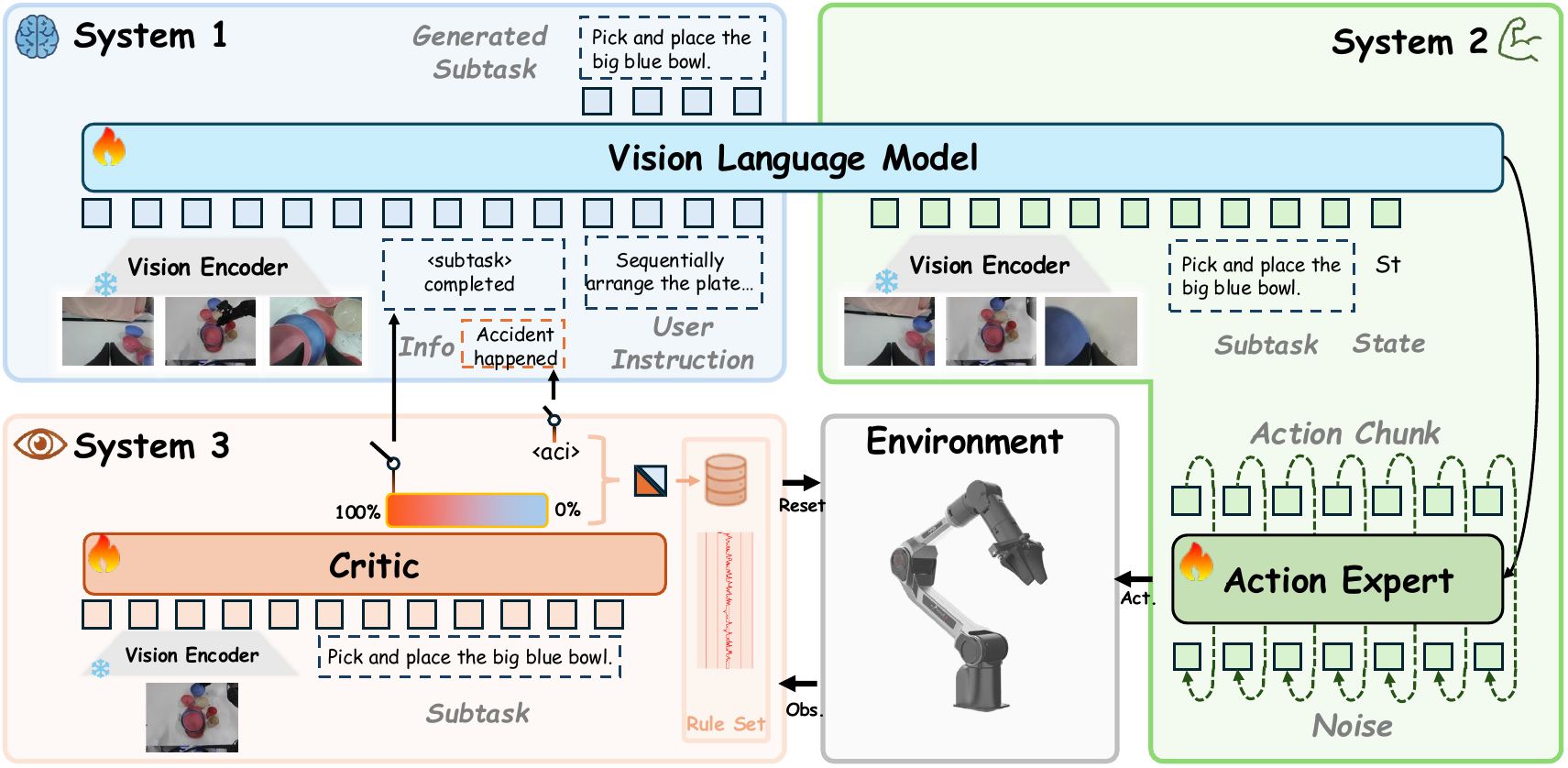}
\end{center}
\caption{Overview of the proposed method. Our Tri-System VLA architecture decouples cognitive reasoning from continuous control via event-driven scheduling. System 2 (Brain) uses a VLM to generate semantic subtasks, while System 1 (Cerebellum) translates them into continuous actions. System 3 (Critic) asynchronously monitors execution, detects anomalies, and integrates human-inspired heuristic rules. By triggering the Brain for replanning only upon completion, failure, or interruption, this asynchronous design effectively bypasses VLM inference bottlenecks in robot control.}
\label{fig:method}
\end{figure}

\section{Methodology}
\label{sec:method}

We propose a \textbf{Tri-System Vision-Language-Action (VLA)} architecture. This section formalizes the manipulation problem, details the Brain-Cerebellum backbone (Systems One and Two), introduces the visually-grounded Critic for state evaluation (System Three), delineates the dynamic scheduling mechanism, and concludes with a scalable, automated data annotation pipeline.

\subsection{Problem Formulation}
\label{subsec:problem}

We formulate long-horizon, language-conditioned manipulation as a dynamic robotic control policy $\pi_\theta$. Unlike traditional fixed-frequency methods, our architecture autonomously toggles between a \textbf{Subtask Generation Mode} (System 2) and an \textbf{Acting Mode} (System 1), strictly governed by an independent \textbf{Critic} $C_\phi$ (System 3). At each timestep $t$, $C_\phi$ evaluates visual observations against the active subtask to assess completion, detect anomalies, and integrate human-inspired heuristic rules, outputting either a progress metric or an emergency switch token (Sec.~\ref{subsec:critic}). If $C_\phi$ indicates task completion or triggers emergency replanning, $\pi_\theta$ enters Subtask Generation Mode. Here, it reasons over current multi-view observations $O_t$, the short-term memory context $m$ from $C_\phi$, and the global instruction $\ell$ to generate a new semantic subtask $g_t \sim \pi_\theta(\cdot \mid O_t, m, \ell)$. Conversely, during nominal execution, $\pi_\theta$ remains in Acting Mode. Conditioned on $O_t$, the active $g_t$, and the robot's proprioceptive state $s_t$, the policy generates precise continuous action chunks $A_t \sim \pi_\theta(\cdot \mid O_t, g_t, s_t)$. Further details on this dual-system execution are provided in Sec.~\ref{subsec:dual_system}.

\subsection{System One and Two: The Brain-Cerebellum Backbone}
\label{subsec:dual_system}

The framework leverages a unified pre-trained Vision-Language Model backbone (e.g., PaliGemma) to embed multi-view visual observations and language into a shared representation space for both the Brain and the Cerebellum.

\subsubsection{System Two (The Brain / Subtask Generation).} The high-level module performs complex cognitive reasoning. It ingests the overarching global user instruction $\ell$ along with an explicit short-term memory context $m$. This memory tracks the immediate execution history, capturing either the successfully completed preceding action (e.g., ``$g_{prev}$ completed'') or a triggered anomaly state (e.g., ``accident happened''). Rather than directly outputting low-level joint torques, the Brain acts as a semantic subgoal generator. Conditioned on the visual observation $O_t$, it autoregressively decodes the current subtask instruction $g_t$ (e.g., ``pick and place the blue cup''). By structuring the prompt context as \textit{``Task: $\ell$; Info: $m$; Current Subtask:''}, we confine the heavy computational bottleneck of the VLM to sparse intervention points while maintaining strict temporal grounding.

\subsubsection{System One (The Cerebellum / Acting).} The low-level module is a dedicated continuous action generation network based on flow matching that bypasses the autoregressive bottleneck entirely. Conditioned on the Brain's currently active semantic subtask $g_t$, the immediate visual observation $O_t$ and robot’s proprioceptive state $s_t$, the Cerebellum learns a vector field $v_t$ to iteratively denoise a random Gaussian distribution into a smooth, deterministic kinematic action chunk $a_{t:t+H}$, where $H$ is the action chunk horizon. By delegating high-frequency, closed-loop control to this flow-matching expert, the system achieves precise, reactive manipulation without being throttled by the Brain's inference latency.

\subsection{System Three: Critic-Guided State Evaluation}
\label{subsec:critic}

To continuously monitor the execution progress of the active subtask $g_t$ and proactively detect anomalies, we introduce a lightweight, visually-grounded \textit{Critic model}, $C_\phi(O_t, g_t)$. 

Unlike traditional critics that require complex auxiliary network heads, we formulate subtask evaluation as a unified Visual Question Answering (VQA) task utilizing a pre-trained Vision-Language Model (e.g., Florence-2). The Critic ingests the visual observation $O_t$ alongside the textual prompt: ``\textit{Evaluate the progress for task: $g_t$}.'' It then autoregressively generates a text string representing either the execution progress or a discrete anomaly state.

\subsubsection{Monte Carlo Value Estimation.} We define the progress value as the normalized expected time-to-completion for the specific subtask. During training, we utilize Monte Carlo estimation derived from trajectory rollouts. For a valid subtask execution of length $L$, the continuous value at step $t$ is formulated as:
$$V_t = \max\left(-1.0, \frac{t - L}{L_{max}^{g_t}}\right)$$
where $L_{max}^{g_t}$ is the robust 90th-percentile maximum length observed for subtask $g_t$ across the training corpus. This mapping normalizes the progress to a strict $[-1.0, 0.0]$ range, where $-1.0$ indicates the start and $0.0$ signifies successful completion. To interface seamlessly with the VLM's text generation paradigm, we discretize $V_t$ into $B = 101$ discrete bins and train the model to output the corresponding bin index as a string.

\subsubsection{Anomaly Detection.} Relying solely on value degradation is often unreliable due to inherent fluctuations (jitter), making it difficult to prevent catastrophic physical failures (e.g., an object dropped). To address this, we replace the numerical bin index with a discrete, high-priority semantic token, \texttt{<aci>}, during the critical temporal window (e.g., the final $20$ frames) of anomalous trajectories. By mapping both progress bins and the \texttt{<aci>} token to the identical text output space, the Critic is trained end-to-end via standard causal language modeling Cross-Entropy loss. This elegant formulation negates the need for separate Binary Cross-Entropy (BCE) anomaly classifiers and mitigates the long-tail effect associated with rare failures, empowering the Critic to emit a hard interrupt signal the moment visual evidence of failure emerges.

\begin{figure}[t]
\begin{center}
\includegraphics[width=\textwidth]{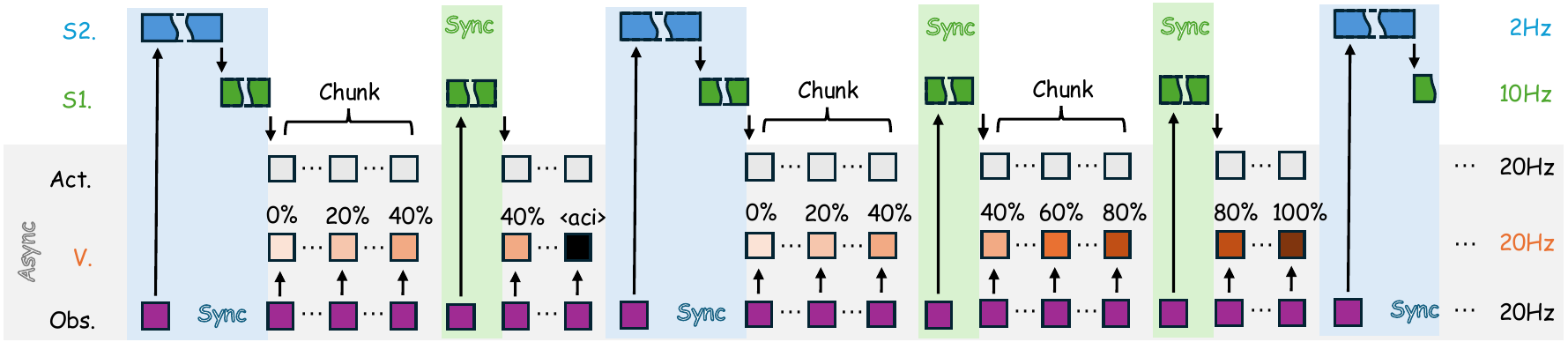}
\end{center}
   \caption{Overview of the Tri-System VLA execution timeline. The System Three Critic ($V.$) asynchronously evaluates progress and governs the dynamic scheduling between the System Two Brain ($S2.$) and the System One Cerebellum ($S1.$).}
\label{fig:timeline}
\end{figure}

\begin{algorithm}[t]
\caption{Critic-Guided Dynamic Scheduling}
\label{alg:async_exec}
\begin{algorithmic}[1]
\REQUIRE User instruction $\ell$, Policy $\pi_{\theta}$, Critic $C_\phi$
\REQUIRE Frequencies: Control $\sim 20$Hz, Max Stagnation $N_{stag}=180$
\STATE Init $A \leftarrow \emptyset$, $t_{stag} \leftarrow 0$, $V_{max} \leftarrow -\infty$, $g_t \leftarrow \pi_{\theta}^{Brain}(O_0, \ell)$
\WHILE{Robot is operational}
    \STATE $O_t, s_t \leftarrow \text{GetObservation()}$
    \STATE $V_t, y_t \leftarrow C_\phi(O_t, g_t)$ \COMMENT{System 3 Async} 
    
    \STATE $t_{stag} \leftarrow (V_t > V_{max}) \text{ ? } 0 \text{ : } t_{stag} + 1$ \COMMENT{Update Critic Tracking}
    \STATE $V_{max} \leftarrow \max(V_{max}, V_t)$
    
    \STATE $m \leftarrow \text{None}$ \COMMENT{Check Preemption Triggers}
    \IF{$y_t == \text{\texttt{<aci>}}$} 
        \STATE $m \leftarrow \text{``accident happened''}$
    \ELSIF{$V_t > \tau_{succ}$} 
        \STATE $m \leftarrow g_t + \text{`` completed''}$
    \ELSIF{$t_{stag} \ge N_{stag}$} 
        \STATE Reset Robot State; $m \leftarrow \text{``stagnation timeout''}$
    \ENDIF
    
    \STATE \textbf{Dynamic Rescheduling:}
    \IF{$m \neq \text{None}$}
        \STATE $A \leftarrow \emptyset$, $V_{max} \leftarrow -\infty$, $t_{stag} \leftarrow 0$
        \STATE $g_t \leftarrow \pi_{\theta}^{Brain}(O_t, \ell, m)$ \COMMENT{System 2 Sync}
    \ENDIF
    
    \IF{$A$ is empty}
        \STATE $A \leftarrow \pi_{\theta}^{cerebellum}(O_t, s_t, g_t)$ \COMMENT{System 1 Sync}
    \ENDIF
    \STATE Execute $a_t \leftarrow A\text{.pop()}$
\ENDWHILE
\end{algorithmic}
\end{algorithm}

\subsection{Dynamic Scheduling}
\label{subsec:async_inference}

Traditional VLA models suffer from inference bottlenecks due to the synchronous coupling of perception, semantic evaluation, and low-level action generation. To circumvent this, we completely decouple these modules using an event-driven, asynchronous scheduling mechanism governed by the System Three Critic. As detailed in Fig.~\ref{fig:timeline}, the architecture operates across three distinct operational cadences:
\begin{itemize}
    \item \textbf{Actuation \& Observation Loop ($\sim 20$ Hz):} To enable real-time monitoring without bottlenecking execution, the Critic asynchronously processes new camera frames $O_t$ in parallel with the robot executing joint commands $a_t$ from the Cerebellum's chunk buffer.
    \item \textbf{System One Sync (Action Generation):} Under nominal conditions, the Cerebellum generates local action chunks conditioned on the active subtask. This synchronization strictly occurs only when the current action buffer is depleted, avoiding redundant kinematic computations.
    \item \textbf{System Two Sync (Subtask Generation):} The computationally heavy Brain is queried on-demand to update the global semantic subtask $g_t$. It remains dormant during nominal execution and is awakened exclusively by Critic-driven interrupts.
\end{itemize}

\subsubsection{Dynamic Preemption Logic.} 
To ensure robust control, the System Three Critic ($C_\phi$) continuously evaluates the current observation $O_t$ conditioned on $g_t$, asynchronously generating a unified text response $y_t$. This text is parsed into either an anomaly flag or de-quantized into an estimated continuous alignment value $V_t \in [-1.0, 0.0]$. As formalized in Algorithm~\ref{alg:async_exec}, the framework preempts the ongoing policy execution upon satisfying any of the following criteria:
\begin{enumerate}
    \item \textbf{Anomaly Detection} ($\text{y}_t == \text{\texttt{<aci>}}$): If the Critic identifies a physical perturbation or execution failure, it instantly flags an accident event.
    \item \textbf{Subtask Completion} ($V_t > \tau_{succ}$): Preemption is triggered when the alignment value surpasses a task-specific success threshold (e.g., $\tau_{succ} \approx -0.041$), facilitating seamless transitions between subtasks.
    \item \textbf{Execution Stagnation} ($t_{stag} \ge N_{stag}$): To prevent policy deadlocks, we introduce a human-inspired heuristic rule. Similar to a human operator aborting a persistently stuck task to better observe the workspace and formulate a new strategy. The system tracks the frames elapsed ($t_{stag}$) since $V_t$ last updated its historical maximum ($V_{max}$), exceeding the stagnation limit triggers both a policy preemption and a physical robot state reset.
\end{enumerate}

Upon triggering any preemption condition, the specific event is encoded as short-term memory context ($m$). The framework immediately flushes the stale action buffer ($A \leftarrow \emptyset$) to prevent incorrect executions and forces an emergency synchronization. The System 2 Brain ($\pi_{\theta}^{Brain}$) is invoked out-of-turn, conditioned on this explicit grounding ($m$), to re-evaluate the visual context and synthesize a corrective sub-goal ($g_t$). Guided by this updated intention, the System 1 Cerebellum ($\pi_{\theta}^{cerebellum}$) subsequently decodes a new, adaptive kinematic trajectory.

\subsection{Automated Subtask Annotation Pipeline}
\label{subsec:annotation}

Training the Tri-System VLA requires datasets with dense, high-level semantic annotations strictly aligned with low-level trajectories. Because manual annotation of subtask boundaries is prohibitively expensive, we propose an automated data annotation pipeline that synergizes physical kinematic heuristics with Vision-Language Model (VLM) retrieval. Our pipeline extracts high-quality semantic demonstrations through two primary stages:

\begin{figure}[t]
\begin{center}
\includegraphics[width=\textwidth]{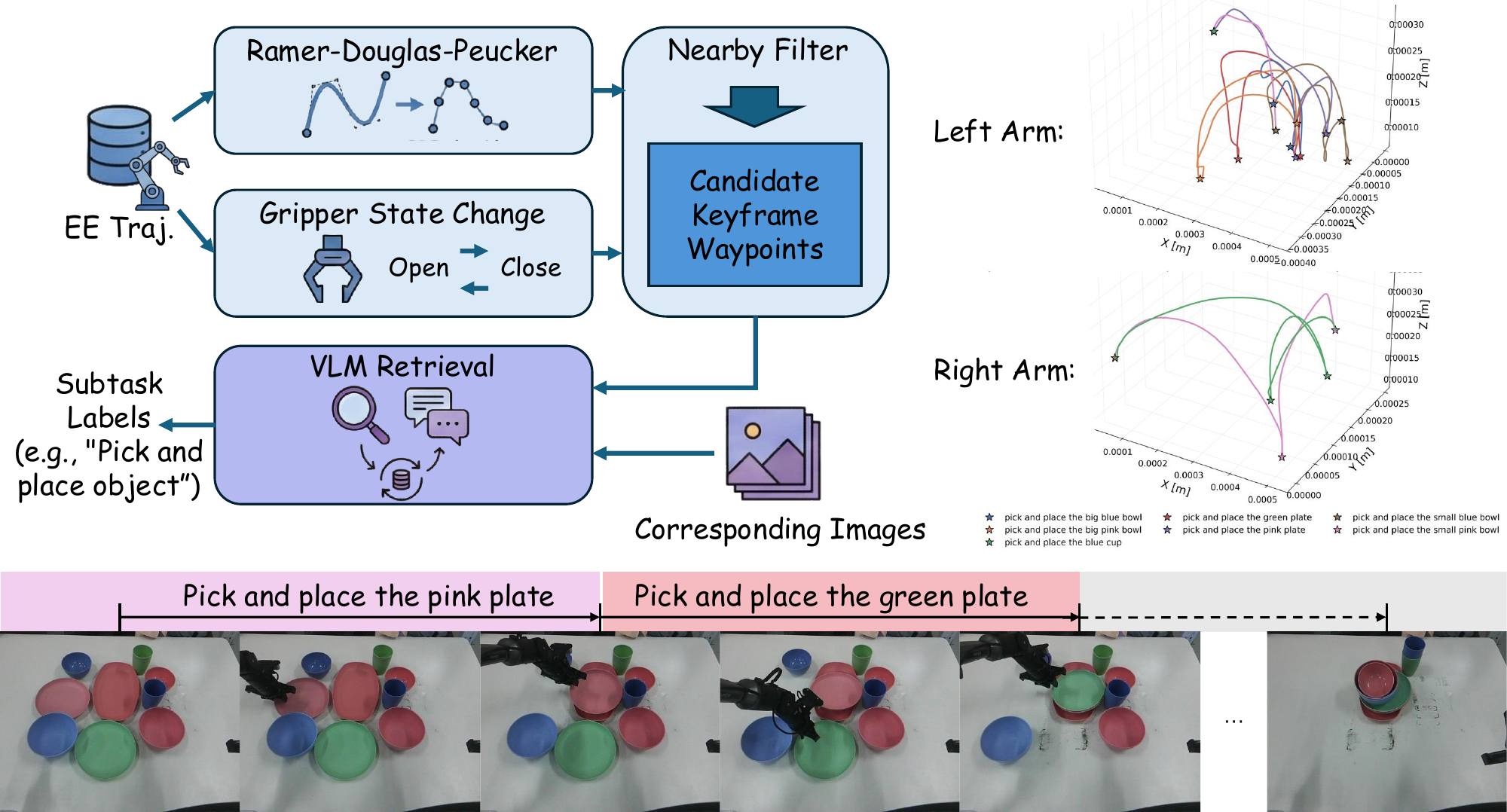}
\end{center}
   \caption{Overview of the automated subtask annotation pipeline. Raw end-effector trajectories are processed into candidate waypoints (top right) via geometric filtering and gripper state analysis. Paired with corresponding visual frames, a VLM retrieves precise semantic labels, resulting in the continuous temporal segmentation and subtask annotation shown at the bottom.}
\label{fig:label}
\end{figure}

\subsubsection{Keyframe Proposal via Kinematics.} We initially parse raw teleoperation trajectories to extract a candidate set of intermediate key states. First, we compute the 3D spatial trajectory of the end-effectors (EE). We then apply the Ramer-Douglas-Peucker (RDP) algorithm \cite{ramer1972iterative} to identify critical geometric waypoints, retaining points $\mathbf{p}_i$ whose orthogonal deviation from the simplified line segment exceeds a distance threshold $\epsilon$:
$$ \max_{i} d_{\perp}(\mathbf{p}_i, \overline{\mathbf{p}_{start} \mathbf{p}_{end}}) > \epsilon $$
These structural waypoints are combined with discrete gripper state changes (e.g., open-to-close transitions indicating a grasp). Finally, we apply a greedy proximity filter that enforces a minimum temporal distance (e.g., $\Delta t \geq 30$ frames) between consecutive keyframes to merge overlapping candidates, yielding a refined set of candidate keyframe waypoints.

\subsubsection{Subtask Grounding via VLM Retrieval.} Relying solely on kinematics inevitably introduces false positives due to operator hesitation or noise. To establish robust semantic boundaries, we extract the corresponding images at the candidate waypoints and pass them into a VLM Retrieval module. The VLM (e.g., Qwen3-VL 32B) queries the visual context against a predefined vocabulary of subtask descriptions to retrieve the most accurate label (e.g., ``Pick and place the pink plate''). Successive candidate waypoints that retrieve identical subtask labels are subsequently merged to form a single, contiguous subtask segment. 

By collaboratively grouping low-level physical waypoints via high-level visual semantic retrieval, this automated pipeline effectively filters out both visual and trajectory noise.

\section{Experiments}
\label{sec:experiments}

To evaluate the effectiveness of our proposed Tri-System VLA, we conduct comprehensive real-world experiments.

\subsection{Experimental Setup}
\label{subsec:exp_setup}

\subsubsection{Hardware Platform.} 
All real-world evaluations are conducted on the Cobot Magic ALOHA platform. This dual-arm robotic system features 7 degrees of freedom (DoF) per arm, enabling dexterous bimanual manipulation. The visual perception suite consists of three Intel RealSense D435 depth cameras: one mounted in a front-facing (head) position to capture the global workspace, and two mounted directly on the left and right wrists to provide localized, ego-centric views during fine-grained execution.

\subsubsection{Implementation Details.} 
We instantiate our Brain-Cerebellum backbone (Systems One and Two) by extending the open-source \texttt{pi0.5} architecture within the \texttt{openpi} framework \cite{pi0.5}. We modify the base \texttt{pi0.5} model to support autoregressive, discrete subtask text generation, serving as our global semantic planner (System Two). The continuous flow-matching action expert (System One) retains the default network capacity and hyperparameters of the original \texttt{pi0.5} implementation to ensure a fair architectural baseline. 

For the Critic (System Three), we employ the lightweight \texttt{Florence-2-base} vision-language model. With approximately 0.2B parameters, it provides the optimal balance of visual reasoning capability and inference speed required for real-time, non-blocking evaluation at 20 Hz. During training, we freeze the vision tower to preserve pre-trained spatial representations and fine-tune only the language and projection layers. The Critic is trained with the AdamW optimizer using a base learning rate of $1 \times 10^{-6}$ following a linear decay schedule with zero warmup steps, for $50$ epochs with a batch size of $64$ per GPU.

\subsubsection{Task Descriptions and Scenarios.}
We evaluate our system on two complex, long-horizon bimanual manipulation tasks:

\textbf{Arrange the Tableware:} The robot is tasked with stacking plates and bowls in order of size, and stacking cups together. To rigorously test robustness, we evaluate this task across four distinct scenarios: 
(1) \textit{Ordered:} Plates, large bowls, small bowls, and cups are placed sequentially from left to right. 
(2) \textit{Scattered:} Large and small bowls are spatially mixed, requiring semantic reasoning over size rather than mere proximity. 
(3) \textit{Left Cup:} The cup is placed out-of-distribution on the left side of the workspace. 
(4) \textit{Fallen:} During execution, a human manually knocks over the cup. The robot must dynamically prioritize righting the cup (at the latest by the end of the current subtask).

\textbf{Tidy up the Desk:} This task involves interacting with deformable objects and consists of four sequential steps: (1) flattening and opening a thin plastic trash bag, (2) picking and placing the first clear plastic bottle into the bag, (3) picking and placing the second bottle into the bag, and (4) picking and placing a crumpled tissue into the bag.

\subsubsection{Data Collection.}
We collected 200 teleoperation trajectories for each task to train the policies. For the \textit{Arrange the Tableware} task, we augmented the dataset with an additional 100 trajectories specifically demonstrating how to recover from a manually knocked-over cup (\textit{Fallen} scenario). Notably, in all collected trajectories for this task, the cups were exclusively positioned on the right side of the workspace, using the right arm, creating an intentional bias to test out-of-distribution generalization. Post-collection, the data was processed using our automated annotation algorithm, followed by manual verification to ensure high-quality subtask boundaries.

\begin{figure}[t]
\begin{center}
\includegraphics[width=\textwidth]{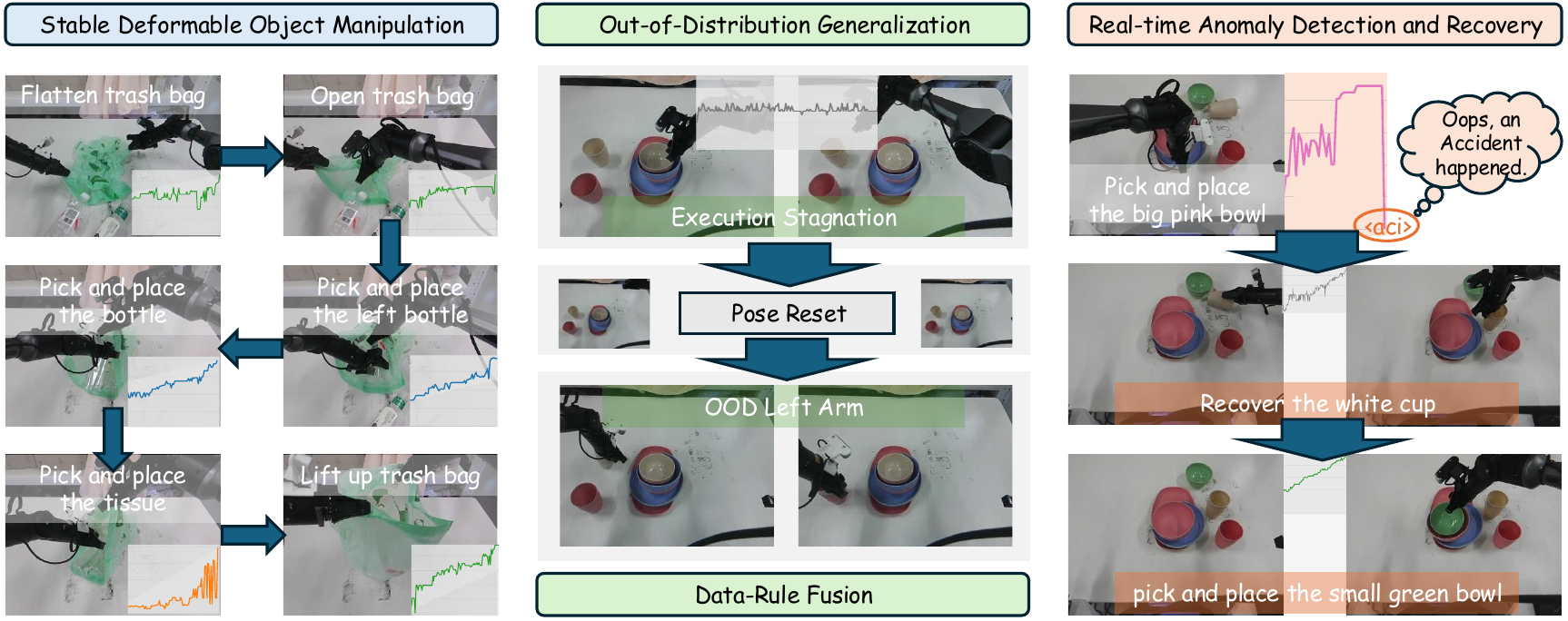}
\end{center}
   \caption{Qualitative results of real-world evaluations. Our proposed system demonstrates robust capabilities across complex scenarios: (Left) stable, long-horizon manipulation of deformable objects; (Middle) out-of-distribution (OOD) generalization via human-inspired rule to resolve execution stagnation; and (Right) real-time anomaly detection (triggered by the <aci> token) followed by autonomous recovery.}
\label{fig:exp_feature}
\end{figure}

\begin{figure}[t]
\begin{center}
\includegraphics[width=\textwidth]{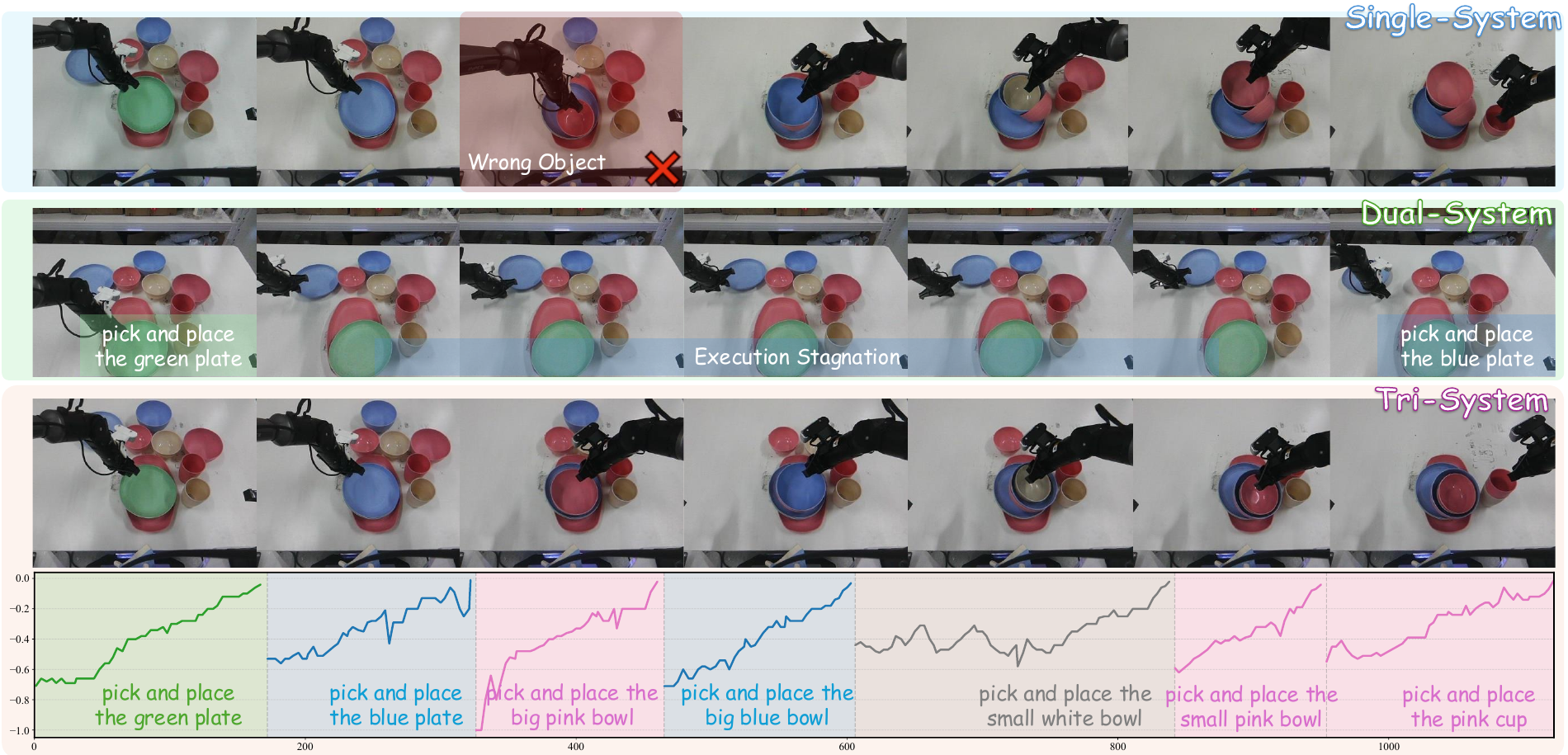}
\end{center}
   \caption{Qualitative comparison on a long-horizon manipulation task. While the Single-System grasps the wrong object and the Dual-System suffers from execution stagnation, our Tri-System successfully completes the continuous multi-step sequence. The bottom plot illustrates the Critic's real-time progress tracking across sequential subtasks.}
\label{fig:exp_compare}
\end{figure}

\subsection{Results and Analysis}
\label{subsec:results}

\subsubsection{Baselines.}
To isolate the contributions of our proposed architecture, we compare against two baselines:
(1) \textbf{Single-System $\pi_{0.5}$:} The standard model predicting actions directly from observations and high-level prompts.
(2) \textbf{Dual-System $\pi_{0.5}$:} A modified model that generates subtasks for \textit{every} action chunk based exclusively on the user prompt, lacking the System Three Critic and short-term memory.

\begin{table}[t]
\caption{Real-World Evaluation Results. We report the number of successful executions over 10 trials for models trained with human data. We separately present the success rates for four scenarios in \textit{Arrange the tableware} and the sequential subtask success rates in \textit{Tidy up the desk}. Bold fonts indicate the best results.}
\centering
\resizebox{\linewidth}{!}{
\begin{tabular}{lcccccccc}
\toprule
\multirow{2}{*}{\textbf{Method}} & \multicolumn{4}{c}{\textbf{Arrange the Tableware}} & \multicolumn{4}{c}{\textbf{Tidy up the Desk}}\\  
\cmidrule(lr){2-5}\cmidrule(lr){6-9}
                        & Ordered  & Scattered  & Left cup & Fallen  & Open & Bottle1 & Bottle2 & Overall \\ \midrule
    Single-System $\pi_{0.5}$ & 8/10 & 0/10 & 0/10 & 2/10 & 7/10 & 5/10 & 2/10 & 0/10  \\
    Dual-System $\pi_{0.5}$ & 7/10 & 6/10 & 1/10 & 5/10 & 6/10 & 5/10 & 1/10 & 0/10 \\
    \textbf{Tri-System $\pi_{0.5}$ (Ours)} & \textbf{10/10} & \textbf{9/10} & \textbf{7/10} & \textbf{7/10} & \textbf{9/10}& \textbf{8/10} & \textbf{5/10} & \textbf{4/10} \\
\bottomrule
\end{tabular}}
\label{tab:main_results}
\end{table}

\subsubsection{Quantitative Results.}
Table \ref{tab:main_results} summarizes the quantitative results. Our proposed Tri-System $\pi_{0.5}$ significantly outperforms both baselines across all evaluated scenarios. 

The \textbf{Single-System $\pi_{0.5}$} demonstrated a fundamental inability to deeply comprehend text conditioning, exhibiting a strong bias toward manipulating the nearest object. Consequently, it completely failed in the \textit{Scattered} scenario (unable to prioritize size over proximity) and the \textit{Fallen} scenario. Furthermore, it severely overfitted the training data distribution; because cups were only manipulated on the right side during training, it exclusively attempted to use its right arm for cups, yielding a 0\% success rate in the \textit{Left Cup} scenario.

While the \textbf{Dual-System $\pi_{0.5}$} improved semantic understanding, its requirement to generate subtasks at every action chunk led to substantial latency. Lacking the Critic's state tracking, it frequently oscillated between different subtasks. This issue was particularly detrimental in \textit{Tidy up the Desk}, where interacting with soft bodies (e.g., the plastic bag) causes continuous visual state changes, causing the Dual-System to rapidly switch intentions and ultimately stall.

Our \textbf{Tri-System $\pi_{0.5}$} utilizes on-demand thinking, which drastically reduces inference latency and prevents task oscillation. The System Three Critic maintains a short-term memory of the current goal, only interrupting System One and triggering System Two for a new plan when a subtask succeeds, an anomaly occurs (e.g., the \textit{Fallen} cup), or the robot stagnates. This allows the robot to handle the \textit{Fallen} scenario natively and smoothly transition through the sequential steps of the \textit{Tidy up the Desk} task without getting trapped in repetitive loops.

\begin{table}[t]
\centering
\caption{Ablation Study. Evaluated on manipulating the out-of-distribution left cup in the ``Arrange the Tableware'' task. We analyze the impact of prompt formulations (\texttt{<arm>} explicitly designates the active manipulator, left or right), prompt sources (System Two vs.\ Ground Truth), and left arm data of bowls (not cup) inclusion.}
\label{tab:ablation_left_cup}
\resizebox{\linewidth}{!}{
\begin{tabular}{llccc}
\toprule
\multirow{2}{*}{\textbf{Method}} & \multirow{2}{*}{\textbf{Prompt Formulation}} & \textbf{Prompt} & \textbf{Left Arm} & \textbf{Success} \\
& & \textbf{Source} & \textbf{Bowl Data} & \textbf{Rate} \\
\midrule
Case 1 & \textit{``pick and place the <obj>''}                   & S2 & $\times$     & 0/10 \\
Case 2 & \textit{``pick and place the <obj>''}                   & S2 & $\checkmark$ & 7/10 \\
Case 3 & \textit{``pick and place the <side> <obj> with <arm>''} & S2 & $\checkmark$ & 3/10 \\
Case 4 & \textit{``pick and place the <side> <obj> with <arm>''} & GT  & $\checkmark$ & \textbf{9/10} \\
\bottomrule
\end{tabular}
}
\end{table}

\subsection{Ablation Study}
\label{subsec:ablation}

To analyze our system's ability to achieve out-of-distribution generalization (e.g., the unseen left-cup task), we conduct a structured analysis of Table \ref{tab:ablation_left_cup}.

\textbf{Q1: What enables OOD execution?} 
The 70\% success rate in Case 2 (vs. 0\% in Case 1) demonstrates that subtask-level training allows System One to learn shared representations. By including left-arm data for other objects (bowls), System One successfully transfers manipulation skills to the OOD cup.

\textbf{Q2: Why do dual-system baselines consistently fail on this task?} 
Baselines lack temporal memory and heuristic logic, suffering from \textit{Execution Stagnation}. After placing a bowl on the right, the right arm remains centered and visually proximal to the left cup. The policy repeatedly attempts unfeasible reaches with the right arm, becoming trapped in a kinematic loop.

\textbf{Q3: How do human-inspired rules resolve this?} 
Our Tri-System uses System Three to detect stagnation and trigger a robot state reset. Retracting the arm breaks the visual-kinematic trap, revealing that the left cup is closer to the left arm and enabling a successful hand-over or replan.

\textbf{Q4: How do prompt formulations impact performance?} 
Case 4's 90\% success rate proves that structured prompts (e.g., \textit{``...with <arm>''}) significantly enhance System One's OOD execution by providing precise spatial-morphological grounding. However, in Case 3 (30\%), using an identical prompt generated by System Two actually degrades performance. This suggests that while System One possesses the latent capacity for OOD tasks, the current System Two (VLM) suffers from distributional overfitting, failing to correctly predict the \textit{<arm>} and \textit{<side>} tokens for unseen scenarios. This identifies System Two's reasoning as the primary bottleneck, suggesting that more powerful VLMs could further bridge this gap.

\section{Conclusions}
In this paper, we presented the Tri-System VLA, an architecture that synergizes high-level reasoning with continuous control via a critic-guided state evaluator (System Three). By decoupling "thinking" from "acting," our framework enables adaptive cognitive switching and autonomous error correction without requiring exhaustive emergency scenario data. Crucially, the architecture allows for the seamless integration of human-inspired heuristic rules, which empowers the system to effectively handle out-of-distribution (OOD) scenarios. Furthermore, our automated subtask annotation pipeline facilitates scalable data synthesis, significantly reducing manual effort. Experiments demonstrate that Tri-System substantially enhances robot robustness and success rates in long-horizon tasks. Future work will focus on integrating Reinforcement Learning (RL) to optimize the model's reasoning proficiency. Additionally, we aim to move beyond reliance on expert demonstrations by leveraging generative world models to synthesize diverse edge-case scenarios, further bolstering the system's generalization in complex environments.

\clearpage  


%
%
\bibliographystyle{splncs04}
\bibliography{main}
\end{document}